\documentclass[11pt,a4paper,reqno]{amsart}
\usepackage{graphicx}
\usepackage{epsfig}
\usepackage{amsmath}
\usepackage{amsfonts}
\usepackage{amssymb}
\usepackage{amscd}
\usepackage{array}
\usepackage{float}
\usepackage{tabularx}
\usepackage{multirow}

\begin{document}

\begin{minipage}[b]{0.5\linewidth}
  {\includegraphics[height=0.82in,width=5.94in]{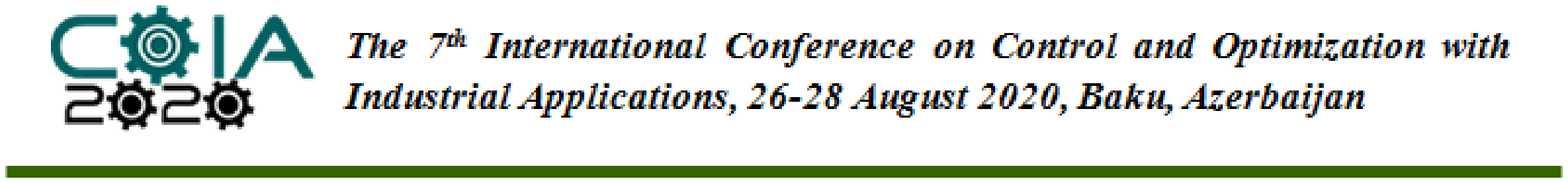}}
\end{minipage}

\bigskip
\title [Pogorilyy S.D.,  Kramov A.A.: Assessment of text coherence based on ... ]
{Assessment of text coherence based on the cohesion estimation}

\author[Pogorilyy S.D.,  Kramov A.A.]  {Pogorilyy S.D.$^1$,  Kramov A.A.$^2$}

\maketitle
\begin{center} $^1$Head of Computer Engineering Department of Faculty of Radio Physics, Electronics and
Computer Systems at Taras Shevchenko National University of Kyiv, Ukraine,
\\ \indent $^2$Computer Engineering Department of Faculty of Radio Physics, Electronics and Computer
Systems at Taras Shevchenko National University of Kyiv, Ukraine,
\\ \indent e-mail: sdp77@i.ua, artemkramovphd@knu.ua \end{center}

\bigskip
\bigskip
\section{Introduction} Coherence evaluation falls into a category of natural language processing (NLP) tasks. The term \textit{"coherence"} implies the semantic integrity of a text. It means that all text components should be logically grouped around a common topic. Such an approach allows information perception in a more convenient way. The assessment of text coherence is widely used in different areas: extraction of texts from heterogeneous pages \cite{1}, copywriting tools, medical notes of patients, etc.

In spite of the need to analyze the semantic similarity between sentences, it is also necessary to take into account the structural connectivity of sentences -  the \textit{cohesion} of a text. The cohesion consists in the availability of common terms, phrases, and objects within a text. Thus, it ensures the lexical and grammatical connectivity of a text that helps an author to address its communication purpose to a reader. As mentioned earlier, the assessment of text coherence should be considered in the field of NLP. Hence, different state-of-the-art methods utilize machine learning techniques (neural networks, Bayesian networks) to solve it. Nevertheless, current methods face the following problems: the neglect of a cohesion component (most methods take into account just the semantic features of a text) and the lack of invisibility of an assessment process.

In this paper, a graph-based coherence estimation method based on the cohesion estimation is suggested. Our method uses a graph-based approach to provide a user with an understanding of the evaluation process. Moreover, it can be applied to different languages, therefore, the effectiveness of this method is examined on the set of English, Chinese and Arabic texts.

\section{Graph-based coherence estimation model}
According to other coherence evaluation methods, a text should be divided into sentences with the further analysis of its connectivity. In order to estimate a cohesion value, it is necessary to analyze common words, entities, and other objects. We suggest to consider all sentences at the level of \textit{phrases}. The successful detection of phrases \cite{2} allows to ensure that the structure of a sentence is correct. Moreover, it prevents from additional removing of stop-words.

Firstly, let consider the cohesion estimation between two sentences $s_{i}$ and $s_{j}$ where $i$ and $j$ represent sentences order correspondingly. Each sentence incorporates the set of tokens:
\begin{equation}
    s_i = \{t_1^i, t_2^i, ..., t_N^i\}\label{eq:1}
\end{equation}
where $N$ - the count of tokens. In order to extract the set of phrases, it is suggested to use the Stanford Open IE approach \cite{3}. This extractor allows retrieving of phrases in the form of relation tuples. Each tuple consists of three elements that incorporate corresponding tokens: subject, relation, and object. Thus, the set of extracted phrases for the sentence $s_{i}$ can be represented in the following way:
\begin{gather}
    P_i = \{P_1^i, P_2^i,....,P_K^i \}\label{eq:4} \\
    P_k^i = Subject_k^i \cup Relation_k^i \cup Object_k^i, k = 1,2,...,K\label{eq:5}
\end{gather}
The next step is to build a bipartite directed graph $B_{ij}=(P_i, P_j, E)$, where $E$ denotes the edges of the graph. The graph $B_{ij}$ should represent the connection between sentences. The weight of edge $e_{lm} \in E$ is calculated in the following way:
\begin{equation}
    e_{lm} = \frac{common(P_l^i, P_m^j)}{unique(P_l^i, P_m^j)}
\end{equation}
where $common(P_l^i, P_m^j)$ is the count of common objects, $unique(P_l^i, P_m^j)$ is the count of unique elements. In case of detection of \textbf{coreferent pairs}, $e_{lm}$ equals to 1. Such an approach allows an increase in the impact of coreference resolution and to take into account long-distance relations (connection between elements can be tracked within a text). It helps to avoid the severe dependence on the language of a text. The lexical similarity of the sentences $sim(s_{i}, s_{j})$ is evaluated as the normalized average weights value of the graph $B_{ij}$:
\begin{equation}
    sim(s_{i}, s_{j}) = \frac{\sum_{e_{lm} \in E}e_{lm}}{|E||i-j|}
\end{equation}

In order to estimate the coherence of a text $T = \{s_1, s_2, ..., s_M\}$, it is necessary to take into account connections between all sentences. Thus, the coherence value of the text should be calculated in the following way:
\begin{equation}
    Coh(T) = \frac{\sum_{i,j \in \{1,2,...,M\}}sim(s_i, s_j)}{M}
\end{equation}

In order to check the effectiveness of the suggested method, two common tasks were solved: document discrimination (DDT) and insertion (IT) tasks. As mentioned earlier, our method does not depend on the features of the certain language. Thus, it was decided to solve these tasks for the following languages: English, Chinese, and Arabic. Such a choice can be explained by the principal differences between these languages because they refer to different language families. The OntoNotes Release 5.0 (LDC2013T19) corpus was used to generate the set of test texts. In order to compare the effectiveness of our method with other graph-based methods, the results obtained in the paper \cite{4} were used.

{\renewcommand{\arraystretch}{1.1}
\begin{table}[H]
\begin{tabularx}{\textwidth}{|l|X|X|X|}
\hline
Language                 & Method               & DDT            & IT \\ \hline
\multirow{4}{*}{English} & SSG                  & 0.774          & \textbf{0.356} \\ \cline{2-4} 
                         & Entity Grid          & 0.845          & 0.346 \\ \cline{2-4} 
                         & Entity Graph         & 0.725          & 0.260 \\ \cline{2-4} 
                         & Cohesion Estimation & \textbf{0.868} & 0.333 \\ \hline
Chinese                  & Cohesion Estimation & 0.760          & 0.343  \\ \hline
Arabic                   & Cohesion Estimation & 0.666          & 0.148  \\ \hline
\end{tabularx}
\caption{\label{tab:first}Accuracy of the method in solving of document discrimination and insertion tasks for different languages.}
\end{table}
}

As can be seen from Table ~\ref{tab:first}, our method (\textbf{Cohesion Estimation}) outperforms other methods \cite{5,6} while solving the document discrimination task. Moreover, it performs sufficient results for Chinese texts in comparison with English corpora. However, it is necessary to focus on Arabic texts because the corresponding tasks were solved with the lowest accuracy among other languages. Such a result can be explained by the low level of terms overlapping within a text due to the inconsistency of terms' spelling in the Arabic language (Arabic diglossia) \cite{7}. The coreference resolution \cite{8} performs the main impact on coherence evaluation for this language. Thus, the semantic models or ontological systems that configured due to the features of the Arabic language should be additionally applied to improve the accuracy of the method.

\section{Conclusions}
In this paper, a new graph-based method of coherence evaluation based on cohesion estimation has been proposed. According to the results obtained the following conclusions can be drawn:
\begin{itemize}
    \item Analysis of a text at the level of phrases allows the additional verification of the consistency of sentences.
    \item Coreference detection should be considered during the coherence evaluation process. Such an approach helps to track the long-distance and neighbor connections between sentences within a text.
    \item The experimental results can indicate that the suggested method can be used to evaluate the coherence of texts for different languages. The accuracy of the method can be increased by the applying of language-dependent instruments of the detection of coreferent objects and common elements. 
\end{itemize}

\bigskip
\bigskip
\noindent \textbf{Keywords:} text coherence, bipartite directed graph, coreference resolution, extraction of phrases, graph-based method of coherence evaluation.

\bigskip
\noindent \textbf{AMS Subject Classification: }68T50.

\end{document}